\documentclass{article}
\usepackage[T1]{fontenc}
\usepackage{origent_preprint,times}

\usepackage{amsmath,amsfonts,bm}

\def\eqref#1{equation~\ref{#1}}

\def\1{\bm{1}}

\DeclareMathAlphabet{\mathsfit}{\encodingdefault}{\sfdefault}{m}{sl}
\SetMathAlphabet{\mathsfit}{bold}{\encodingdefault}{\sfdefault}{bx}{n}

\usepackage{xcolor}
\definecolor{linkblue}{RGB}{24,78,128}
\usepackage{xurl}
\usepackage[colorlinks=true,linkcolor=linkblue,citecolor=linkblue,urlcolor=linkblue]{hyperref}
\usepackage{xspace}
\usepackage{graphicx}
\usepackage{capt-of}
\usepackage{booktabs}
\usepackage{array}
\usepackage{longtable}
\usepackage{placeins}

\newcommand{\origent}{OriGent\xspace}
\newcommand{\codeurl}{\href{https://github.com/Henry-Jessie/origent}{\texttt{github.com/\allowbreak Henry-Jessie/\allowbreak origent}}}
\newcommand{\projecturl}{\href{https://henry-jessie.github.io/origent}{\texttt{henry-jessie.github.io/\allowbreak origent}}}
\newcommand{\hfurl}{\href{https://huggingface.co/henryen}{\texttt{huggingface.co/\allowbreak henryen}}}
\newcommand{\headericon}[1]{\raisebox{-0.22ex}{\includegraphics[height=1.08em]{#1}}}

\title{\origent: Post-Training Hardware Engineering Agents through Scalable Environment Synthesis}

\author{Fan Cui, Kexing Zhou, Youwei Zhuo \\
Peking University \\
\texttt{pku\_cf@stu.pku.edu.cn}, \texttt{zhoukexing@pku.edu.cn}, \texttt{youwei@pku.edu.cn} \\[5pt]
{\small
\href{https://huggingface.co/henryen}{\headericon{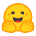}\hspace{0.36em}\textbf{Hugging Face}}\hspace{1.4em}%
\href{https://github.com/Henry-Jessie/origent}{\headericon{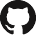}\hspace{0.36em}\textbf{GitHub}}\hspace{1.4em}%
\href{https://henry-jessie.github.io/origent}{\headericon{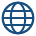}\hspace{0.36em}\textbf{Website}}%
}
}

\iclrfinalcopy
\begin{document}

\maketitle

\begin{abstract}
Post-training coding agents relies on large-scale executable environments, yet building such environments for hardware engineering remains difficult.
A single fix often spans register-transfer-level (RTL) circuits, build configurations, and verification code, all checked through repository-specific simulation and electronic design automation (EDA) toolchains; as a result, repository-level interaction data stays scarce.
We introduce \origent, an environment-synthesis and post-training framework that turns merged pull requests from public hardware repositories into executable training tasks.
For each candidate change, \origent restores the pre-fix repository state, builds and validates tests against the reference repair, and drafts a problem statement grounded in those verified checks.
Separating reusable toolchain profiles from per-PR task setup produces 18,257 executable tasks across 90 repositories covering RTL design, verification, EDA tools, hardware DSLs, and firmware.
Using these environments, we post-train Qwen3.6-35B-A3B with long-context supervised fine-tuning on verified DeepSeek-V4.1 Flash trajectories, followed by synchronous online GRPO on 2,163 difficulty-screened tasks from disjoint repositories.
On HWE-Bench v1.1 (417 tasks), the resolve rate rises from 40.8\% for the base model to 66.8\% after SFT and 74.2\% after RL (3-run mean), a gain of 33.4 percentage points.
Code, sanitized tasks and trajectories, deduplicated RL container images, and model weights are released at \codeurl{} and \hfurl{} (project page: \projecturl{}).
\end{abstract}

\begin{figure}[!htb]
    \centering
    \includegraphics[width=\linewidth]{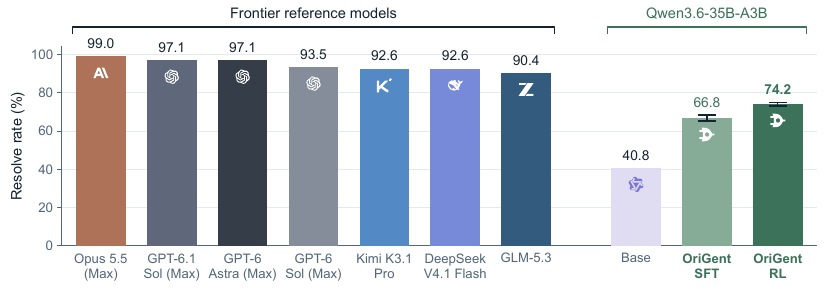}
    \caption{\textbf{HWE-Bench v1.1 performance.} Overall resolve rates on 417 tasks. Error bars on \origent-SFT and \origent-RL show the min--max range across three independent evaluation runs.}
    \label{fig:hwe_overview}
\end{figure}

\clearpage

\section{Introduction}
\label{sec:introduction}

Executable training environments have enabled coding agents to resolve real issues across large software repositories~\citep{pan2024swegym,yang2025swesmith}.
In hardware engineering, a single bug fix frequently touches register-transfer-level (RTL) circuit descriptions, build configurations, and verification code, all checked through electronic design automation (EDA) tools and cycle-accurate simulators.
A controller, for instance, can compile cleanly and still fail simulation because a handshake fires one clock cycle late.

To handle such bugs, an agent needs hands-on experience with how code edits and tool invocations change build and simulation outcomes.
Prior domain-specific training has improved standalone Verilog generation and comprehension~\citep{cui2024origen,liu2025deeprtl}, and simulation feedback has been used to train testbench revision models~\citep{zhang2026llm4cov}.
Scaling repository-level hardware training environments, however, faces three obstacles.
First, public hardware code is scarce~\citep{liu2025deeprtl}: Verilog and SystemVerilog make up roughly 0.10\% of files in the training subset of The Stack v3~\citep{lozhkov2026stackv3}, and few repositories maintain curated issue trackers paired with standalone reproduction tests.
Second, each project depends on its own mix of simulator versions, HDL libraries, testbenches, and build flags.
Third, each task needs a problem statement aligned with what its tests actually check, so wrong patches fail while valid alternative implementations still pass.

We address these bottlenecks with \origent, an environment-synthesis and post-training framework for repository-level hardware agents.
Rather than depending on issue trackers and pre-existing tests, \origent builds executable tasks directly from merged pull requests (PRs).
Construction agents restore the base commit, author target tests that fail before the reference fix and pass afterward alongside regression tests that pass on both states, and draft a problem statement grounded in those verified checks.
Because tasks within the same repository share most of their build stack, \origent factors out reusable toolchain profiles from per-PR task setup.
Applied to 90 public repositories across RTL design, verification, EDA software, hardware DSLs, and firmware, the pipeline produces 18,257 executable tasks.

These environments serve both supervised fine-tuning (SFT) and online reinforcement learning (RL).
Starting from Qwen3.6-35B-A3B~\citep{qwen2026qwen36}, we first perform long-context SFT on verified DeepSeek-V4.1 Flash~\citep{deepseek2026v41flash} trajectories, then run synchronous online GRPO~\citep{shao2024deepseekmath} on 2,163 difficulty-screened tasks from 23 disjoint repositories.
On HWE-Bench v1.1~\citep{cui2026hwebench} (417 tasks across six unseen repositories), the resolve rate climbs from 40.8\% for the base model to 66.8\% after SFT and 74.2\% after RL (3-run mean; Figure~\ref{fig:hwe_overview}).
Over 51 RL updates, held-out HWE-Bench accuracy rises alongside steady gains in same-task training pass@1 across epochs and longer test-time debugging trajectories.

Our main contributions are:
\begin{itemize}
    \item \textbf{Scalable hardware environment synthesis.} A PR-to-task pipeline that separates reusable repository toolchain profiles from per-task test construction and aligns problem statements with execution-validated checks.
    \item \textbf{Open hardware agent training resources.} A 90-repository synthesis pool of 18,257 tasks, together with an open release of 148 repository toolchain profiles, 11,024 sanitized Harbor tasks, 8,447 sanitized teacher trajectories, and 2,159 deduplicated RL OCI images.
    \item \textbf{Open-weight hardware engineering agents.} \origent-SFT and \origent-RL, which raise HWE-Bench v1.1 resolve rate by 26.0 and 33.4 percentage points over the base model.
\end{itemize}

\section{Related Work}
\label{sec:related_work}

\paragraph{Hardware engineering models and benchmarks.}
Most hardware post-training work targets standalone RTL generation or narrow verification subtasks.
RTLCoder~\citep{liu2025rtlcoder}, CraftRTL~\citep{liu2024craftrtl}, and DeepRTL~\citep{liu2025deeprtl} train models on synthetic RTL generation and comprehension data, while CodeV-R1~\citep{zhu2025codevr1} and ChipMATE~\citep{yu2026chipmate} add execution feedback via reinforcement learning for RTL and reference-model generation.
On the verification side, PRO-V-R1~\citep{zhao2025provr1}, CovR~\citep{abdelatty2026covr}, and LLM4Cov~\citep{zhang2026llm4cov} use functional or coverage signals to train testbench generators and revisers.
Evaluation has progressed from module-level Verilog benchmarks~\citep{liu2023verilogeval,lu2024rtllm} to broader design and verification suites~\citep{pinckney2025cvdp,yu2026chipbench,wang2026rtlbenchls}, and most recently to HWE-Bench~\citep{cui2026hwebench}, which tests historical bug repair inside project-native build and simulation flows.
Open post-training pipelines for hardware, however, have not kept pace with this shift to full-repository tasks.

\begin{table}[!t]
    \caption{\textbf{Comparison of hardware post-training works and repository-level environment-synthesis pipelines.}}
    \label{tab:training_resource_comparison}
    \centering
    \footnotesize
    \setlength{\tabcolsep}{3pt}
    \renewcommand{\arraystretch}{1.15}
    \begin{tabular}{@{}>{\raggedright\arraybackslash}p{0.225\linewidth}
        >{\raggedright\arraybackslash}p{0.15\linewidth}
        >{\raggedright\arraybackslash}p{0.185\linewidth}
        >{\raggedright\arraybackslash}p{0.28\linewidth}
        >{\raggedright\arraybackslash}p{0.10\linewidth}@{}}
        \toprule
        Work & Domain & Environment setup & Tests \& statements & Training \\
        \midrule
        DeepRTL\newline\citep{liu2025deeprtl}
        & Hardware (module)
        & None
        & Synthetic RTL \& comprehension specs
        & SFT \\[3pt]
        CodeV-R1\newline\citep{zhu2025codevr1}
        & Hardware (module)
        & Per-module simulator
        & Unit testbenches \& specs
        & RL \\[3pt]
        LLM4Cov\newline\citep{zhang2026llm4cov}
        & Hardware (testbench)
        & Per-module simulator
        & Coverage-guided TB revisions
        & SFT \\
        \midrule
        SWE-Gym\newline\citep{pan2024swegym}
        & Software (repo)
        & Per-issue container
        & Existing tests \& issue text
        & SFT \\[3pt]
        SWE-smith\newline\citep{yang2025swesmith}
        & Software (repo)
        & 1 env / repo snapshot
        & Injected bugs \& existing tests
        & SFT \\[3pt]
        R2E-Gym\newline\citep{jain2025r2egym}
        & Software (repo)
        & Per-commit container
        & Synthesized tests \& statements
        & SFT \\[3pt]
        CodeMidas\newline\citep{ye2026codemidas}
        & Software (repo)
        & Per-repo container
        & Synth.\ tests \& specs from code
        & RL \\[3pt]
        SWE-rebench V2\newline\citep{badertdinov2026swerebenchv2}
        & Software (repo)
        & 1 env / repo snapshot
        & Existing tests \& filtered issues
        & Data \\[3pt]
        Scale-SWE\newline\citep{zhao2026scaleswe}
        & Software (repo)
        & Nearest-PR env reuse
        & Existing/synth.\ tests \& stmts
        & SFT \\
        \midrule
        \textbf{\origent\ (Ours)}
        & \textbf{Hardware (repo)}
        & \textbf{Multi-era profiles}
        & \textbf{Synth.\ target/reg.\ \& reviewed stmts}
        & \textbf{SFT + RL} \\
        \bottomrule
    \end{tabular}
\end{table}

\paragraph{Executable training environments.}
In software engineering, SWE-Gym~\citep{pan2024swegym} recovers containers for historical issues, while R2E-Gym~\citep{jain2025r2egym} and Scale-SWE~\citep{zhao2026scaleswe} synthesize tests and problem statements directly from commits and pull requests.
Alternative pipelines generate tasks without historical bugs: SWE-smith~\citep{yang2025swesmith} injects defects checked by existing tests, and CodeMidas~\citep{ye2026codemidas} derives tests from existing repository functionality.
Closer to our reusable-profile design, SWE-rebench V2~\citep{badertdinov2026swerebenchv2} amortizes installation and test setup at the repository level across multilingual software projects.
Outside general software, similar executable gyms have been built for biomedical coding~\citep{xu2026medagentgym}, scientific tool use~\citep{shen2026sciagentgym}, and SQL debugging~\citep{li2025swesql}.
Hardware repositories break three assumptions common to these pipelines: a single repository snapshot rarely builds older revisions across coupled simulator and EDA tool versions, more than half of substantive PRs modify no test files, and raw issue descriptions often omit or leak tested behavior.
\origent addresses all three by separating multi-era repository toolchain profiles from per-PR task setup, synthesizing focused target and regression tests verified in fresh containers, and rewriting problem statements with an independent review pass.

\paragraph{Agent post-training.}
Recent work trains repository-level coding agents through trajectory SFT with selective error masking~\citep{tao2026swelego}, SFT followed by online RL on executable software tasks~\citep{song2026swemaster}, and RL directly inside native agent harnesses~\citep{du2026legorl,yang2026t1,ivison2026tmax}.
When long debugging sessions exceed the context window, CompactionRL~\citep{li2026compactionrl} and SUPO~\citep{lu2026supo} jointly optimize interaction turns and context summaries under task-level rewards.
Hardware debugging episodes routinely span hundreds of turns and multiple context compactions across slow simulation containers.
Building on these ideas, \origent pairs failure-preserving turn masking and standalone compaction supervision in SFT with native-harness, 256K-context online GRPO, using token-in/token-out rollout capture, rollout routing replay~\citep{ma2026routingreplay}, and frozen MoE routers to keep long-horizon MoE updates stable.
Table~\ref{tab:training_resource_comparison} summarizes how \origent compares with prior hardware and software training pipelines.

\section{Scalable Environment Synthesis}
\label{sec:environment_synthesis}

\begin{figure}[!t]
    \centering
    \includegraphics[width=0.92\linewidth]{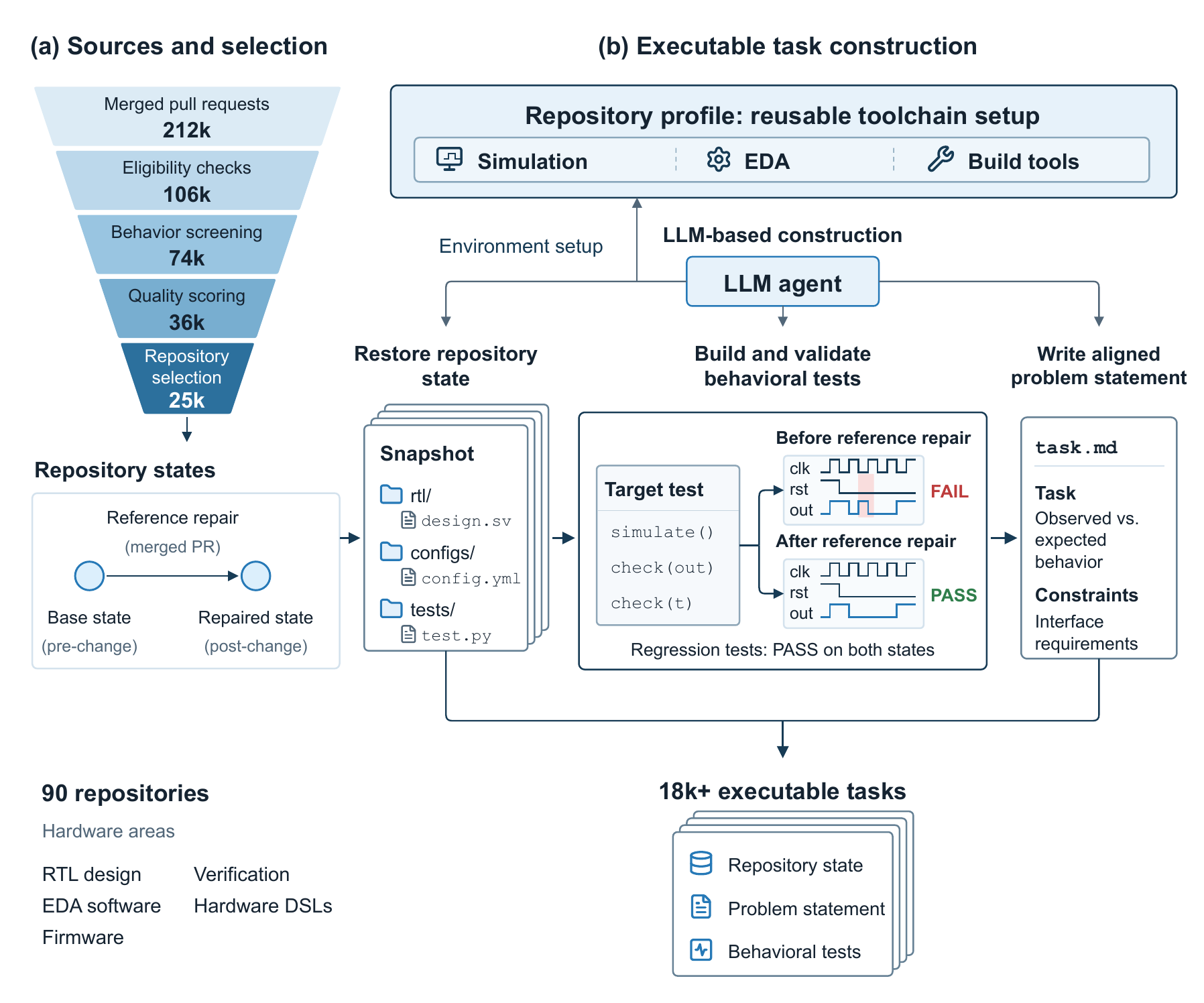}
    \caption{\textbf{Environment synthesis in \origent.} \textbf{(a)}~Merged pull requests from hardware repositories are screened to select candidate changes, each providing a base state and a reference repair. \textbf{(b)}~Using reusable toolchain environments, construction agents restore repository states and validate tests before generating problem statements aligned with the tested behavior. The resulting executable tasks feed post-training (Figure~\ref{fig:origent-training}).}
    \label{fig:origent-framework}
\end{figure}

\origent converts merged pull requests from public hardware repositories into self-contained executable tasks, each comprising a pre-built task container image, a problem statement, and target and regression test suites (Figure~\ref{fig:origent-framework}).

\subsection{Repository and Environment Preparation}
\label{sec:environment_preparation}

We collect merged pull requests from 150 public hardware repositories discovered by combining curated seed projects with GitHub topic and language queries, excluding the six HWE-Bench evaluation repositories at discovery time.
After filtering by merge date and patch size, an LLM screen removes changes dominated by documentation, maintenance chores, or generated code.
Each remaining candidate is scored from 0 to 2 on five dimensions from its diff, PR description, and linked discussions: \emph{task value} (substantive bug fix or behavioral change), \emph{behavior clarity}, \emph{test signal} (externally checkable behavior, even without existing tests), \emph{scope complexity}, and \emph{interface clarity}.
We keep candidates that score 2 on task value and at least 1 on every other dimension, excluding changes that require physical hardware to verify.
Retaining repositories with at least 20 qualifying PRs and deferring Zephyr so that a single embedded OS project does not dominate the pool leaves 25,003 candidates across 90 repositories (Figure~\ref{fig:origent-framework}a; Appendix~\ref{app:candidate_filtering}).

Because tasks from the same repository share most of their build stack, environment setup is split into two stages.
First, a \emph{profile agent} inspects the repository's build files and defines reusable \emph{repository profiles} specifying base images, compilers, simulators, and libraries.
Distinct historical eras within one project can use separate profiles: SpinalHDL, for example, uses separate profiles for legacy sbt/JDK~8 and modern sbt/mill/JDK~17 toolchains.
Each profile image is validated on a small batch of candidate tasks before broad reuse.
Second, for each PR, a \emph{task agent} selects a profile image, checks out the base commit, configures task-specific build setup, and flags any missing shared dependencies for profile updates.
To prevent evaluation or training agents from recovering the future fix from Git history, we strip the original \texttt{.git} metadata, initialize a clean repository at the base commit, and snapshot the resulting container as the task's prepared image.

\subsection{Test Construction and Problem Statement Generation}
\label{sec:behavioral_synthesis}

Over half of our candidate PRs modify no test or verification files, and full repository regression suites often take too long to simulate inside an online RL loop.
The task agent therefore synthesizes two focused suites per PR against the reference diff and surrounding codebase: \emph{target tests} (including minimal testbenches or stimuli when needed) that check the intended change, and \emph{regression tests} that guard related functionality.
Iterating on live compiler and simulator output, the agent is instructed to test observable behavior rather than private implementation details; exact diagnostic strings or internal symbol names are asserted only when mandated by the PR discussion or a public interface.

A task passes validation only when all four checks succeed in fresh containers launched from the task's prepared image: target tests must fail on the base state with a designated behavioral assertion (rejecting build or dependency errors and timeouts) and pass once the reference patch is applied, while regression tests must pass in both states.
On our SFT tasks, passing solver patches edit 72.9\% of reference source files on average while reusing a median of only 11.7\% of reference added lines, confirming that validated tests accept alternative implementations rather than memorizing the reference diff (Appendix~\ref{app:construction_agents}).

Raw hardware PR descriptions frequently omit interface details exercised by the tests or reveal the fix itself.
After test validation, a generation agent drafts a self-contained problem statement from the original issue or PR, the base code, and the verified tests, specifying all tested public interfaces and expected behavior without exposing repair hints.
Because the generation agent sees the reference patch, a separate review agent independently cross-checks the draft against the source PR and execution logs to supply missing tested constraints, resolve ambiguities, and strip leaked implementation details; this review pass revised 51.7\% of drafted statements (Appendix~\ref{app:construction_agents}).
Appendix~\ref{app:task_example} walks through a complete constructed task.

\subsection{Dataset Overview}
\label{sec:task_coverage}

\begin{figure}[!t]
    \centering
    \includegraphics[width=\linewidth]{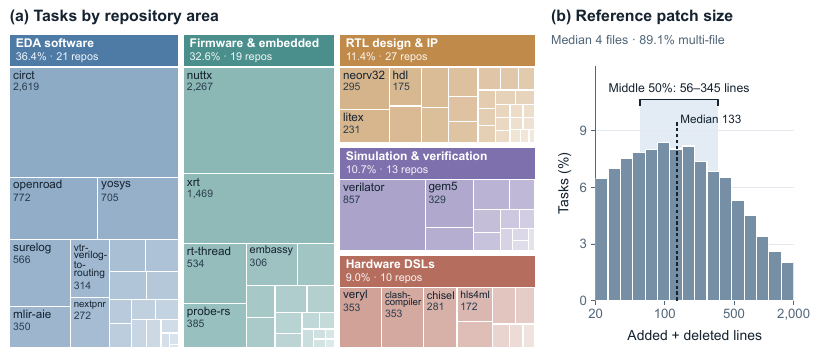}
    \caption{\textbf{Dataset overview.} \textbf{(a)}~Tasks by repository area; each tile is one of the 90 repositories, sized by its task count. \textbf{(b)}~Reference patch sizes (added plus deleted lines) on a log scale; the shaded band marks the interquartile range.}
    \label{fig:dataset_overview}
\end{figure}

The pipeline yields 18,257 executable tasks across 90 repositories built on 152 toolchain profiles (Appendix~\ref{app:construction_agents}).
EDA software (36.4\%) and firmware and embedded software (32.6\%) make up the two largest shares, with the remaining tasks split across RTL design and IP, simulation and verification, and hardware DSLs and generators (Figure~\ref{fig:dataset_overview}a).

Reference patches modify a median of 133 lines across four files (Figure~\ref{fig:dataset_overview}b).
Most fixes span multiple files (89.1\%), and 36.6\% touch multiple languages or intermediate representations, rising to 61.8\% in EDA software and 56.4\% in simulation and verification.
Compared with software tasks from SWE-rebench V2.1 and Scale-SWE, our SFT-pool reference patches are roughly twice as large in changed lines, and a controlled GLM-5.3 solver in OpenCode requires more than twice as many turns and generated tokens to solve them (Appendix~\ref{app:swerebench_comparison}).

For post-training and public release, we partition the synthesis pool into disjoint SFT and RL repositories: cleaning the SFT split retains 8,861 tasks, while difficulty screening across 23 RL repositories narrows 8,199 candidates to 2,163 informative tasks (Section~\ref{sec:online_rl}), giving 11,024 sanitized Harbor-format tasks in total.
Appendix~\ref{app:dataset_composition} lists all 90 repositories with their SFT and RL counts.

\section{Post-Training Hardware Engineering Agents}
\label{sec:post_training}

\begin{figure}[!t]
    \centering
    \includegraphics[width=0.78\linewidth]{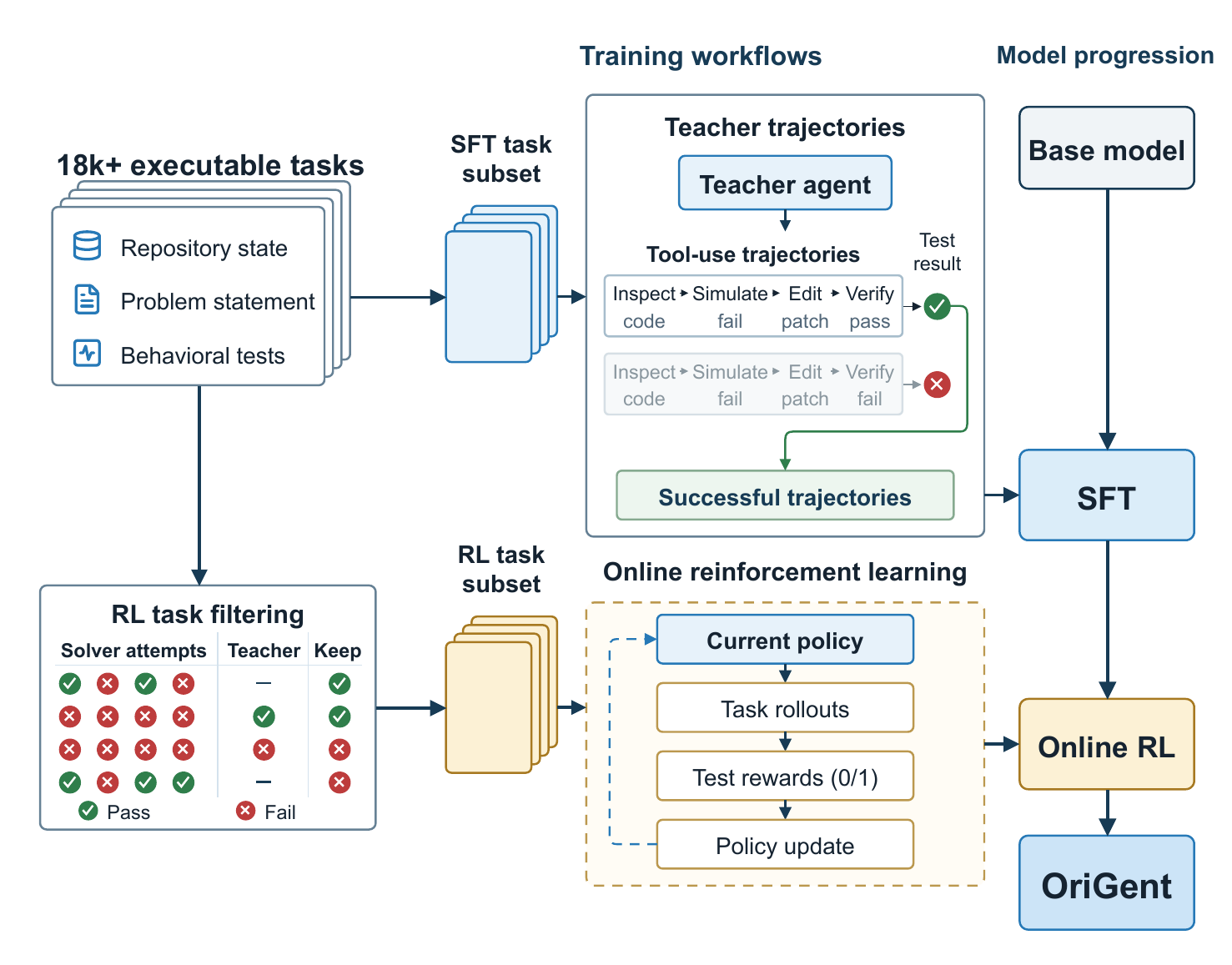}
    \caption{\textbf{Post-training in \origent.} The executable tasks are split into an SFT subset and an RL subset. The model undergoes supervised fine-tuning on successful teacher trajectories, followed by online reinforcement learning on filtered tasks with test-based rewards.}
    \label{fig:origent-training}
\end{figure}

We post-train Qwen3.6-35B-A3B~\citep{qwen2026qwen36} in two stages, long-context SFT followed by synchronous online RL (Figure~\ref{fig:origent-training}), using OpenCode~\citep{anomaly2026opencode} as the agent harness throughout.
The SFT, RL, and HWE-Bench evaluation repositories are mutually disjoint.

\subsection{Supervised Fine-Tuning}
\label{sec:sft}

We collect teacher demonstrations by running DeepSeek-V4.1 Flash~\citep{deepseek2026v41flash} on the SFT-pool tasks inside no-network containers.
We retain attempts whose final changes pass both target and regression tests, excluding trajectories with more than 300 assistant turns, two or more context compactions, external network access attempts, or known task defects.
This leaves 8,447 verified trajectories across 61 repositories.

During SFT, a turn-level loss mask zeroes the loss on selected assistant turns in full (reasoning, visible text, and tool calls) while keeping those turns and their tool outputs in context.
To curb verbose reasoning loops and malformed calls, we mask incomplete responses, framework-reported tool errors, turns with empty reasoning, and turns whose reasoning exceeds 8,192 tokens; turns that trigger compiler or simulation failures remain unmasked so the model learns to diagnose errors and recover.
Because OpenCode generates each compaction summary in a separate call outside the main turn history, we also reconstruct standalone summary-generation examples at the exact prefix where compaction was triggered (segment counts and filtering details in Appendix~\ref{app:sft_statistics}).

\subsection{Online Reinforcement Learning}
\label{sec:online_rl}

Online RL initializes from \origent-SFT and optimizes the policy with synchronous Group Relative Policy Optimization (GRPO)~\citep{shao2024deepseekmath}.
To focus rollout compute on tasks with informative reward variance, we screen candidates from the RL repositories with up to four Qwen3.6-27B~\citep{qwen2026qwen3627b} attempts each, keeping tasks solved once or twice alongside zero-pass tasks that GLM-5.3~\citep{zai2026glm53} solves in the same prepared image.
This yields 2,163 RL tasks across 23 repositories (Appendix~\ref{app:rl_selection_counts}).

In each update, the policy samples a group of rollouts per task and receives a binary reward for passing both target and regression tests.
Advantages are mean-centered within each group without standard-deviation scaling, KL and entropy penalties are set to zero, and each batch takes a single optimizer step.
Zero-variance groups (all-pass or all-fail) and infrastructure-aborted groups are dropped dynamically, with additional tasks sampled until the target batch of mixed-outcome groups is filled~\citep{yu2025dapo}.

\medskip
\noindent\begin{minipage}[t]{0.51\textwidth}
Long debugging sessions can reach OpenCode's working-context threshold before finishing.
When triggered, the agent summarizes its progress and resumes from the summary with the modified workspace intact (Figure~\ref{fig:context_compaction}).
Because downstream actions depend on this summary, admitted summaries are supervised jointly with interaction turns under the episode's binary reward~\citep{li2026compactionrl,lu2026supo}.
Compaction splits a trajectory into segments that share its group advantage, with token losses normalized across all supervised tokens in the prompt group.
\end{minipage}\hfill
\begin{minipage}[t]{0.46\textwidth}
    \vspace{0pt}
    \centering
    \includegraphics[width=\linewidth]{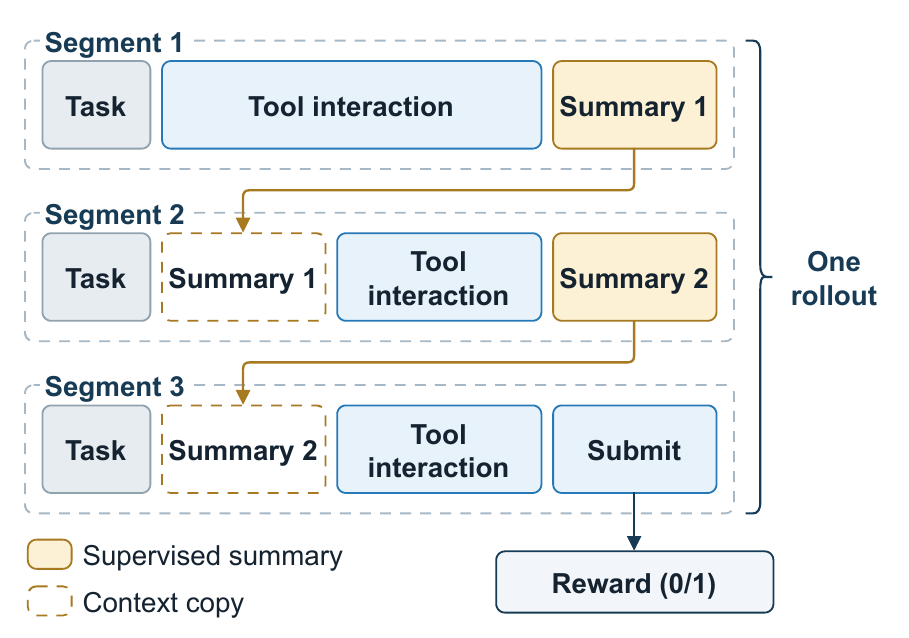}
    \captionof{figure}{\textbf{Context compaction.} Shaded summaries receive policy supervision, whereas dashed copies only supply context; all segments share the episode reward.}
    \label{fig:context_compaction}
\end{minipage}
\medskip

Long-horizon MoE training is sensitive to tokenization drift and expert-routing mismatch between inference and training.
We keep rollout and training aligned via token-in/token-out (TITO) capture in Miles~\citep{miles2026}, rollout routing replay ($R^3$)~\citep{ma2026routingreplay}, and frozen MoE router weights.
We further mask length-truncated and n-gram-repetitive turns on passing trajectories while keeping negative gradients on failed ones, and discard rare groups corrupted by inference-side NaN logits (Appendix~\ref{app:training_configuration}).

\section{Experiments}
\label{sec:experiments}

\subsection{Experimental Setup}
\label{sec:experimental_setup}

\paragraph{Benchmark and models.}
We evaluate on HWE-Bench v1.1~\citep{cui2026hwebench}, which contains 417 executable hardware bug-repair tasks across six held-out open-source repositories: Ibex (35), CVA6 (35), Caliptra (16), Rocket Chip (32), XiangShan (54), and OpenTitan (245).
We compare Qwen3.6-35B-A3B~\citep{qwen2026qwen36} before post-training (Base), after SFT (\origent-SFT), and after online RL (\origent-RL), alongside the seven frontier reference systems on the HWE-Bench v1.1 leaderboard: Claude Opus 5.5~\citep{anthropic2026opus55}, GPT-6.1 Sol~\citep{openai2026gpt61}, GPT-6 Astra and GPT-6 Sol~\citep{openai2026gpt6}, Kimi K3.1 Pro~\citep{moonshot2026kimi31}, DeepSeek-V4.1 Flash~\citep{deepseek2026v41flash}, and GLM-5.3~\citep{zai2026glm53}.

\paragraph{Training and evaluation protocol.}
SFT trains for 3 epochs (906 updates) with cosine decay from $10^{-5}$ to $10^{-6}$, and online GRPO runs for 51 updates at a constant learning rate of $10^{-6}$ starting from \origent-SFT (step~0) to produce \origent-RL (step~51; full hyperparameters in Appendix~\ref{app:training_evaluation_settings}).
Base, \origent-SFT, and \origent-RL are evaluated in OpenCode v1.18.22 with a 256K context window, 64K max output tokens per turn, temperature 1.0, top-$p$ 1.0 (0.95 on the 50 prompt-revised Base tasks), context compaction with 72K reserved tokens (tool-output pruning disabled), and a 6-hour per-task wall-clock limit.
Frontier models are reported from the official HWE-Bench v1.1 leaderboard in their respective native harnesses (Claude Code, Codex CLI, Kimi Code, DeepSeek Harness, and OpenCode).
To account for sampling variance on long-horizon tasks, we evaluate \origent-SFT and 17 RL checkpoints across the 51-update run, repeating all 417 tasks three times per checkpoint and reporting 3-run means with min--max ranges.

\subsection{Main Results on HWE-Bench v1.1}
\label{sec:main_results}

\paragraph{Overall and per-repository resolve rates.}
Table~\ref{tab:hwe_main_results} summarizes performance across HWE-Bench v1.1.
Long-context SFT raises Qwen3.6-35B-A3B from 40.8\% to 66.8\% (+26.0 percentage points), and online GRPO reaches 74.2\% (+7.4 points, or +33.4 points overall).

\begin{table}[!htb]
    \caption{\textbf{HWE-Bench v1.1 results.} Resolve rates (\%) by repository and overall across 417 tasks. \origent-SFT and \origent-RL report 3-run means (runs: 272/285/279 and 311/312/305, respectively).}
    \label{tab:hwe_main_results}
    \centering
    \small
    \setlength{\tabcolsep}{3pt}
    \begin{tabular*}{\linewidth}{@{\extracolsep{\fill}}lrrrrrrr@{}}
        \toprule
        Model & Ibex & CVA6 & Caliptra & Rocket & XiangShan & OpenTitan & Overall \\
        Tasks & 35 & 35 & 16 & 32 & 54 & 245 & 417 \\
        \midrule
        \multicolumn{8}{l}{\textit{Frontier reference models}} \\
        Claude Opus 5.5 (Max) & \textbf{100.0} & \textbf{100.0} & \textbf{100.0} & \textbf{100.0} & \textbf{100.0} & 98.4 & \textbf{99.0} \\
        GPT-6.1 Sol (Max) & 97.1 & 97.1 & \textbf{100.0} & 87.5 & 96.3 & 98.4 & 97.1 \\
        GPT-6 Astra (Max) & 97.1 & 97.1 & \textbf{100.0} & 84.4 & 96.3 & \textbf{98.8} & 97.1 \\
        GPT-6 Sol (Max) & 97.1 & 97.1 & \textbf{100.0} & 84.4 & 92.6 & 93.5 & 93.5 \\
        Kimi K3.1 Pro & 88.6 & 94.3 & \textbf{100.0} & 87.5 & 87.0 & 94.3 & 92.6 \\
        DeepSeek-V4.1 Flash & 85.7 & 97.1 & \textbf{100.0} & 87.5 & 87.0 & 94.3 & 92.6 \\
        GLM-5.3 & 77.1 & 94.3 & \textbf{100.0} & 90.6 & 83.3 & 92.7 & 90.4 \\
        \midrule
        \multicolumn{8}{l}{\textit{Qwen3.6-35B-A3B (OpenCode v1.18.22)}} \\
        Qwen3.6-35B-A3B (Base) & 40.0 & 68.6 & 43.8 & 34.4 & 31.5 & 39.6 & 40.8 \\
        \origent-SFT & 70.5 & \textbf{85.7} & \textbf{87.5} & 65.6 & 66.0 & 62.6 & 66.8 \\
        \textbf{\origent-RL} & \textbf{77.1} & \textbf{85.7} & 85.4 & \textbf{70.8} & \textbf{71.6} & \textbf{72.4} & \textbf{74.2} \\
        \bottomrule
    \end{tabular*}
\end{table}

The two post-training stages play complementary roles across repositories.
SFT produces broad improvements on every project because the base model frequently terminates before completing repository-specific build and simulation workflows.
Once SFT establishes reliable tool invocation and test iteration, online RL concentrates its gains on complex SoC and processor codebases where SFT leaves the most headroom, led by OpenTitan (+9.8 points), Ibex (+6.6), XiangShan (+5.6), and Rocket Chip (+5.2), whose bugs often cut across coupled RTL modules, verification harnesses, or firmware.
By contrast, CVA6 holds at 85.7\% and the 16-task Caliptra subset remains within sampling noise of its SFT score (difference of 0.3 solved tasks on average).
Together, SFT and online RL close more than half of the base model's gap to proprietary and frontier leaderboard systems (90.4\%--99.0\% in their native harnesses), establishing a strong open-weight baseline on repository-level hardware repair.

\begin{figure}[!t]
    \centering
    \includegraphics[width=\linewidth]{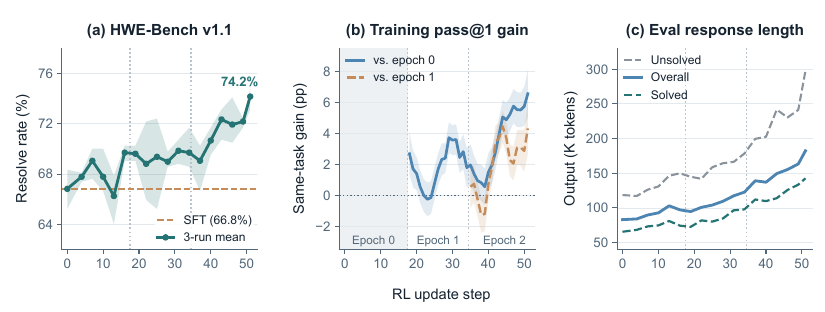}
    \caption{\textbf{Online RL progression across 51 GRPO updates.} \textbf{(a)}~HWE-Bench v1.1 resolve rate of \origent-SFT (step~0) and 17 RL checkpoints (3 runs per checkpoint; shaded band shows min--max range). \textbf{(b)}~Same-task training pass@1 gain relative to earlier epochs (5-rollout pooled window, shaded $\pm 1$ standard error). \textbf{(c)}~Mean evaluation output tokens on HWE-Bench v1.1 separated by task outcome. Dotted vertical lines mark the start of training epochs 1 and 2 (steps 18 and 35).}
    \label{fig:rl_training_dynamics}
\end{figure}

\paragraph{Online RL progression.}
Figure~\ref{fig:rl_training_dynamics} tracks evaluation and training dynamics across all 51 updates.
Held-out HWE-Bench v1.1 resolve rate climbs steadily over the first two training epochs before accelerating in the third epoch to reach 74.2\% at step~51 (Figure~\ref{fig:rl_training_dynamics}a).
Because training tasks are reshuffled each epoch, raw per-step reward mixes policy improvement with batch difficulty.
Comparing each task directly against its own pass@1 in earlier epochs removes this confound (Figure~\ref{fig:rl_training_dynamics}b): same-task pass@1 reaches +6.6 percentage points over epoch~0 and +4.3 points over epoch~1 by step~51, rising monotonically from 65.7\% to 69.2\% on tasks seen in all three epochs.
Throughout training, policy entropy decays smoothly without collapse and train--rollout divergence stays tight (median steady-state KL 0.0012), confirming that TITO, $R^3$, and frozen MoE routers keep multi-hour updates numerically on-policy (Appendix~\ref{app:train_rollout_consistency}).

\subsection{Agent Behavior across Post-Training}
\label{sec:behavioral_analysis}

Beyond resolve rates, we examine how SFT and online RL reshape interaction depth, context management, and test-time compute on HWE-Bench v1.1 (Table~\ref{tab:behavior_scaling}).

\begin{table}[!htb]
    \caption{\textbf{Evaluation behavior on HWE-Bench v1.1 across post-training stages.} Per-task interaction and generation statistics across all 417 held-out tasks in OpenCode v1.18.22 (single run for Base; 3-run means for \origent-SFT and \origent-RL). Assistant turns count agent interaction steps excluding context-compaction summaries.}
    \label{tab:behavior_scaling}
    \centering
    \small
    \begin{tabular*}{\linewidth}{@{\extracolsep{\fill}}lrrr@{}}
        \toprule
        Metric & Base & \origent-SFT & \origent-RL \\
        \midrule
        Assistant turns per task (mean) & 24.5 & 125.2 & 222.4 \\
        Assistant turns per task (median) & 20.0 & 71.3 & 120.8 \\
        Tool calls per task (mean) & 31.0 & 158.9 & 313.8 \\
        Context compactions per task (mean) & 0.01 & 0.93 & 2.71 \\
        Output tokens, overall mean (K) & 19.7 & 83.2 & 182.3 \\
        Output tokens, overall median (K) & 11.1 & 45.9 & 82.8 \\
        Output tokens, solved tasks mean (K) & 11.6 & 65.7 & 142.7 \\
        Output tokens, unsolved tasks mean (K) & 25.4 & 118.6 & 297.6 \\
        \bottomrule
    \end{tabular*}
\end{table}

\paragraph{From brief attempts to multi-compaction debugging.}
Post-training expands interaction horizons in two distinct steps.
Base Qwen3.6-35B-A3B averages 24.5 assistant turns and almost never reaches the compaction threshold, frequently stopping after an initial edit or failed build.
Long-context SFT increases interaction turns and tool calls fivefold, turning multi-cycle compilation, simulation, and log inspection into default behavior and making context compaction routine on longer repairs.
Because online GRPO optimizes binary task resolution without a length penalty, \origent-RL nearly doubles interaction turns and triples compactions again relative to \origent-SFT; filtered RL training batches show the same steady expansion from the first five to the last five updates, rising from 240.8 to 293.2 mean turns and from 1.86 to 3.08 compactions per trajectory (Appendix~\ref{app:train_rollout_consistency}).
At these horizons, supervising compaction summaries directly during SFT and RL is exercised multiple times per hard task rather than only in rare outliers.

\paragraph{Productive iteration versus wall-clock exhaustion.}
Separating evaluation trajectories by outcome (Figure~\ref{fig:rl_training_dynamics}c and Table~\ref{tab:behavior_scaling}) clarifies where the added compute goes.
On solved tasks, mean output grows from 11.6K tokens for Base to 65.7K for \origent-SFT and 142.7K for \origent-RL as the model persists through deeper fault-localization and regression cycles on complex repositories.
Unsolved tasks expand even more sharply, reaching 297.6K tokens under \origent-RL and driving the wide gap between overall mean and median output before hitting the 6-hour wall-clock limit.
Extra test-time search is therefore central to the +7.4-point RL gain, even though a substantial share of the added budget goes into attempts that still fail.

\section{Conclusion}
\label{sec:conclusion}

We presented \origent, an environment-synthesis and post-training framework for repository-level hardware engineering agents.
Decoupling reusable repository toolchain profiles from per-PR test construction and grounding problem statements in verified execution checks turns public hardware pull requests into reliable training environments at scale.
Across post-training, long-context SFT equips the base model with repository-specific build and simulation workflows, while synchronous online RL with compaction-aware supervision and on-policy MoE replay drives deeper multi-turn fault localization on complex hardware codebases, lifting Qwen3.6-35B-A3B from 40.8\% to 74.2\% on HWE-Bench v1.1.

\paragraph{Open-source artifacts.}
We release the synthesis, post-training, and evaluation codebase (including 90 repository templates defining 148 toolchain profiles) under Apache-2.0 at \codeurl{}, together with \origent-SFT and \origent-RL weights, sanitized SFT trajectories and Harbor tasks, and deduplicated RL container images at \hfurl{} (project page: \projecturl{}).

\bibliography{iclr2027_conference}
\bibliographystyle{iclr2027_conference}

\appendix
\clearpage

\section{Dataset Collection and Analysis}
\label{app:dataset_collection_analysis}

\subsection{Candidate Collection and Filtering}
\label{app:candidate_filtering}

We collected 257,099 pull requests (PRs) from 150 public GitHub repositories by combining 83 seed hardware projects with GitHub topic and language queries covering \texttt{riscv}, \texttt{fpga}, \texttt{rtl}, \texttt{cocotb}, \texttt{Verilog}, and \texttt{SystemVerilog}.
Queries excluded forks and archived repositories and applied star thresholds ($>$300 for \texttt{riscv}, \texttt{fpga}, and \texttt{Verilog}; $>$100 for \texttt{rtl}; $>$80 for \texttt{SystemVerilog}; $>$50 for \texttt{cocotb}).
HWE-Bench's six repositories were removed prior to PR collection.
Two initial query batches contributed 55 and 34 repositories (the second requiring at least 100 historical PRs), and a follow-up search added 61 repositories spanning HDL tools, FPGA toolchains, firmware, and hardware emulators.

Table~\ref{tab:candidate_filtering} summarizes each filtering step.
We treat the full PR diff as the reference change; linked issues and existing tests are optional.
GLM-5.2 first filters out PRs limited to documentation, build or maintenance chores, generated code, or purely structural refactors.
GPT-5.5 then scores each remaining candidate from 0 to 2 along five criteria: \emph{task value} (2 for substantive bug fixes or behavior changes such as state-machine, bus-protocol, or register-logic repairs), \emph{behavior clarity}, \emph{test signal} (checkable observable behavior even without existing tests), \emph{scope complexity} (non-trivial logic edits), and \emph{interface clarity}.
We retain candidates scoring 2 on task value and at least 1 on each remaining criterion, while dropping PRs tagged as maintenance, vendor/generated updates, repository syncs, unclear interfaces, or physical-hardware-only checks.

Keeping repositories with at least 20 qualifying PRs leaves 35,173 candidates across 91 repositories.
Deferring Zephyr (10,170 candidates) leaves 25,003 PRs across 90 repositories for environment construction.

\begin{table}[!htb]
    \caption{\textbf{Candidate filtering funnel.} Criteria and retained PR counts across filtering stages.}
    \label{tab:candidate_filtering}
    \centering
    \small
    \begin{tabular*}{\linewidth}{@{\extracolsep{\fill}}lp{0.50\linewidth}r@{}}
        \toprule
        Stage & Filtering criterion & Retained PRs \\
        \midrule
        Merge status & Merged pull requests only & 211,829 \\
        Merge date & Merged between 1 Jan 2017 and 29 May 2026 & 205,563 \\
        Patch availability & Retrieved, nonempty reference diff & 205,203 \\
        Patch size & 20--2,000 added + deleted lines across $\leq$50 files & 106,251 \\
        Behavior screening & Behavioral hardware or software changes & 74,034 \\
        Quality scoring & Task value 2; other scores $\geq 1$; no exclusion tags & 35,504 (134 repos) \\
        Repository threshold & At least 20 qualifying candidates per repository & 35,173 (91 repos) \\
        Construction scope & Defer Zephyr candidates & 25,003 (90 repos) \\
        \bottomrule
    \end{tabular*}
\end{table}

\subsection{Dataset Characteristics}
\label{app:dataset_composition}

The synthesized collection contains 18,257 tasks from 90 repositories.
From this pool, we release 11,024 sanitized Harbor tasks: 8,861 in the SFT pool across 62 repositories, from which trajectory filtering retains 8,444 training tasks across 61 repositories, and 2,163 in the disjoint RL pool across 23 repositories.

Tasks are grouped by the primary engineering role of their source repository.
Simulation and verification covers HDL simulators, verification frameworks, and architecture or instruction-set emulators.
Firmware and embedded software also includes device runtimes, drivers, and host-side embedded development tools.

\paragraph{Changes across files and languages.}
Patch size counts added plus deleted lines across the full reference diff (including tests, configuration, documentation, comments, and blank lines); binary files count toward file totals but not line totals.
Across the 18,257 tasks, 89.1\% of reference patches edit at least two files (84.6\% when restricted to recognized code, script, and IR extensions).
Grouping files by language extension, 36.6\% of patches touch at least two languages or representations, led by EDA software (61.8\%) and simulation and verification (56.4\%), compared with 6.1\% in firmware and embedded software.

\begin{figure}[!htb]
    \centering
    \includegraphics[width=\linewidth]{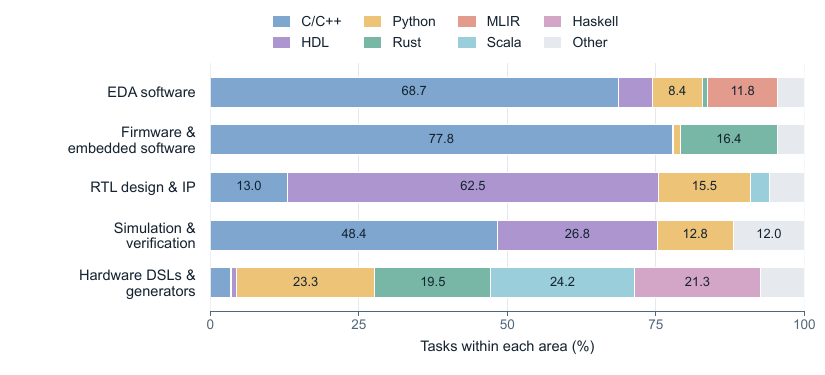}
    \caption{\textbf{Dominant language of reference patches by repository area.} Percentages are normalized within each area. HDL combines Verilog, SystemVerilog, and VHDL. Other includes remaining languages, ties, and patches without a recognized code extension.}
    \label{fig:language_by_area}
\end{figure}

\paragraph{Languages across engineering areas.}
Figure~\ref{fig:language_by_area} breaks patches down by dominant language (the recognized extension with the most added plus deleted lines).
C/C++ dominates firmware and embedded software (77.8\%) and EDA software (68.7\%), alongside Rust in firmware (16.4\%) and MLIR in EDA (11.8\%).
Hardware DSLs and generators split more evenly across Scala (24.2\%), Python (23.3\%), Haskell (21.3\%), and Rust (19.5\%).

Table~\ref{tab:all_repositories} lists all 90 source repositories and the number of tasks from each that enter SFT and RL.
CIRCT, NuttX, and XRT together account for 34.8\% of all tasks.

\begingroup
\small
\setlength{\tabcolsep}{5pt}
\setlength{\LTleft}{0pt}
\setlength{\LTright}{0pt}
\setlength{\LTcapwidth}{\linewidth}
\begin{longtable}{@{\extracolsep{\fill}}p{0.58\linewidth}lrrr@{}}
\caption{\textbf{Complete repository inventory.} Unique task counts in the synthesized collection (All tasks) and in the SFT and RL training sets. Area labels denote RTL design and IP (RTL), simulation and verification (Sim.), EDA software (EDA), hardware DSLs and generators (DSL), and firmware and embedded software (FW).}\label{tab:all_repositories}\\
\toprule
Repository & Area & All tasks & SFT & RL \\
\midrule
\endfirsthead
\multicolumn{5}{l}{\tablename\ \thetable\ (continued)}\\
\toprule
Repository & Area & All tasks & SFT & RL \\
\midrule
\endhead
\midrule
\multicolumn{5}{r}{Continued on next page}\\
\endfoot
\bottomrule
\endlastfoot
\nolinkurl{amaranth-lang/amaranth} & DSL & 128 & 0 & 29 \\
\nolinkurl{analogdevicesinc/hdl} & RTL & 175 & 0 & 35 \\
\nolinkurl{antonblanchard/microwatt} & RTL & 43 & 40 & 0 \\
\nolinkurl{apache/nuttx} & FW & 2,267 & 0 & 616 \\
\nolinkurl{bespoke-silicon-group/basejump_stl} & RTL & 74 & 70 & 0 \\
\nolinkurl{black-parrot/black-parrot} & RTL & 163 & 0 & 46 \\
\nolinkurl{cartesi/machine-emulator} & Sim. & 20 & 17 & 0 \\
\nolinkurl{chipsalliance/caliptra-mcu-sw} & FW & 221 & 0 & 41 \\
\nolinkurl{chipsalliance/chisel} & DSL & 281 & 207 & 0 \\
\nolinkurl{chipsalliance/i3c-core} & RTL & 22 & 19 & 0 \\
\nolinkurl{chipsalliance/surelog} & EDA & 566 & 527 & 0 \\
\nolinkurl{chipsalliance/uhdm} & EDA & 147 & 142 & 0 \\
\nolinkurl{chipsalliance/verible} & EDA & 64 & 0 & 0 \\
\nolinkurl{clash-lang/clash-compiler} & DSL & 353 & 0 & 143 \\
\nolinkurl{cocotb/cocotb} & Sim. & 152 & 0 & 36 \\
\nolinkurl{cornell-zhang/allo} & DSL & 46 & 44 & 0 \\
\nolinkurl{earlephilhower/arduino-pico} & FW & 120 & 0 & 36 \\
\nolinkurl{efabless/openlane2} & EDA & 17 & 0 & 0 \\
\nolinkurl{embassy-rs/embassy} & FW & 306 & 0 & 59 \\
\nolinkurl{enjoy-digital/litedram} & RTL & 45 & 39 & 0 \\
\nolinkurl{enjoy-digital/liteeth} & RTL & 22 & 14 & 0 \\
\nolinkurl{enjoy-digital/litex} & RTL & 231 & 0 & 48 \\
\nolinkurl{esp-rs/esp-hal} & FW & 41 & 0 & 2 \\
\nolinkurl{f4pga/f4pga-arch-defs} & EDA & 126 & 0 & 41 \\
\nolinkurl{f4pga/prjxray} & EDA & 58 & 49 & 0 \\
\nolinkurl{fastmachinelearning/hls4ml} & DSL & 172 & 148 & 0 \\
\nolinkurl{ferrandi/panda-bambu} & EDA & 4 & 3 & 0 \\
\nolinkurl{fpga-research/fabulous} & EDA & 33 & 25 & 0 \\
\nolinkurl{freertos/freertos-kernel} & FW & 121 & 100 & 0 \\
\nolinkurl{gem5/gem5} & Sim. & 329 & 0 & 122 \\
\nolinkurl{ghdl/ghdl} & Sim. & 37 & 36 & 0 \\
\nolinkurl{glasgowembedded/glasgow} & FW & 15 & 9 & 0 \\
\nolinkurl{google/xls} & EDA & 74 & 0 & 0 \\
\nolinkurl{greatscottgadgets/luna} & RTL & 20 & 18 & 0 \\
\nolinkurl{intel/rohd} & DSL & 98 & 95 & 0 \\
\nolinkurl{lekkit/rvvm} & Sim. & 23 & 16 & 0 \\
\nolinkurl{limine-bootloader/limine} & FW & 34 & 24 & 0 \\
\nolinkurl{llvm/circt} & EDA & 2,619 & 2,310 & 0 \\
\nolinkurl{lnis-uofu/openfpga} & RTL & 107 & 92 & 0 \\
\nolinkurl{logisim-evolution/logisim-evolution} & Sim. & 105 & 0 & 30 \\
\nolinkurl{lowrisc/sonata-system} & RTL & 40 & 0 & 0 \\
\nolinkurl{mikepopoloski/slang} & EDA & 145 & 135 & 0 \\
\nolinkurl{modm-io/modm} & FW & 80 & 0 & 0 \\
\nolinkurl{najaeda/naja} & EDA & 66 & 60 & 0 \\
\nolinkurl{openhwgroup/core-v-verif} & Sim. & 89 & 73 & 0 \\
\nolinkurl{openhwgroup/cv-hpdcache} & RTL & 24 & 23 & 0 \\
\nolinkurl{openhwgroup/cv32e40p} & RTL & 87 & 78 & 0 \\
\nolinkurl{openhwgroup/cv32e40s} & RTL & 58 & 56 & 0 \\
\nolinkurl{openhwgroup/cv32e40x} & RTL & 134 & 120 & 0 \\
\nolinkurl{openhwgroup/cvw} & RTL & 123 & 101 & 0 \\
\nolinkurl{oreboot/oreboot} & FW & 29 & 23 & 0 \\
\nolinkurl{pavel-demin/red-pitaya-notes} & RTL & 28 & 23 & 0 \\
\nolinkurl{probe-rs/probe-rs} & FW & 385 & 0 & 100 \\
\nolinkurl{pulp-platform/ara} & RTL & 53 & 41 & 0 \\
\nolinkurl{pulp-platform/axi} & RTL & 38 & 32 & 0 \\
\nolinkurl{pulp-platform/cheshire} & RTL & 20 & 12 & 0 \\
\nolinkurl{pulp-platform/common_cells} & RTL & 26 & 21 & 0 \\
\nolinkurl{pulp-platform/floonoc} & RTL & 15 & 13 & 0 \\
\nolinkurl{pymtl/pymtl3} & DSL & 54 & 49 & 0 \\
\nolinkurl{riscv-boom/riscv-boom} & RTL & 11 & 8 & 0 \\
\nolinkurl{riscv-collab/riscv-openocd} & FW & 102 & 79 & 0 \\
\nolinkurl{riscv-non-isa/riscv-arch-test} & Sim. & 52 & 45 & 0 \\
\nolinkurl{riscv-software-src/opensbi} & FW & 18 & 17 & 0 \\
\nolinkurl{riscv-software-src/riscv-isa-sim} & Sim. & 155 & 140 & 0 \\
\nolinkurl{rt-thread/rt-thread} & FW & 534 & 0 & 132 \\
\nolinkurl{rustsbi/rustsbi} & FW & 10 & 6 & 0 \\
\nolinkurl{siliconcompiler/siliconcompiler} & EDA & 163 & 0 & 29 \\
\nolinkurl{slaclab/surf} & RTL & 153 & 0 & 48 \\
\nolinkurl{spinalhdl/spinalhdl} & DSL & 118 & 0 & 25 \\
\nolinkurl{stnolting/neorv32} & RTL & 295 & 0 & 83 \\
\nolinkurl{sysprog21/rv32emu} & Sim. & 70 & 55 & 0 \\
\nolinkurl{the-openroad-project/openlane} & EDA & 14 & 0 & 0 \\
\nolinkurl{the-openroad-project/openroad} & EDA & 772 & 706 & 0 \\
\nolinkurl{trabucayre/openfpgaloader} & FW & 12 & 11 & 0 \\
\nolinkurl{ucb-bar/chipyard} & RTL & 42 & 31 & 0 \\
\nolinkurl{ucb-bar/gemmini} & RTL & 24 & 22 & 0 \\
\nolinkurl{ucsbarchlab/pyrtl} & DSL & 39 & 35 & 0 \\
\nolinkurl{verilator/verilator} & Sim. & 857 & 0 & 181 \\
\nolinkurl{verilog-to-routing/vtr-verilog-to-routing} & EDA & 314 & 298 & 0 \\
\nolinkurl{veryl-lang/veryl} & DSL & 353 & 324 & 0 \\
\nolinkurl{vhdl-ls/rust_hdl} & EDA & 60 & 21 & 0 \\
\nolinkurl{vunit/vunit} & Sim. & 42 & 42 & 0 \\
\nolinkurl{xilinx/mlir-aie} & EDA & 350 & 320 & 0 \\
\nolinkurl{xilinx/pynq} & FW & 53 & 47 & 0 \\
\nolinkurl{xilinx/xrt} & FW & 1,469 & 1,162 & 0 \\
\nolinkurl{yosyshq/apicula} & EDA & 73 & 61 & 0 \\
\nolinkurl{yosyshq/nextpnr} & EDA & 272 & 0 & 120 \\
\nolinkurl{yosyshq/sby} & Sim. & 26 & 24 & 0 \\
\nolinkurl{yosyshq/yosys} & EDA & 705 & 0 & 161 \\
\nolinkurl{zigembeddedgroup/microzig} & FW & 126 & 86 & 0 \\
\midrule
Total & & 18,257 & 8,444 & 2,163 \\
\end{longtable}
\endgroup

\subsection{Comparison with Software Environment-Synthesis Pipelines}
\label{app:swerebench_comparison}

To compare task complexity against representative software environment-synthesis pipelines, Table~\ref{tab:swerebench_comparison} contrasts our SFT-pool tasks with SWE-rebench V2.1~\citep{badertdinov2026swerebenchv2} and Scale-SWE~\citep{zhao2026scaleswe} under a single controlled reference solver (GLM-5.3 in OpenCode).
Reference pull-request sizes count all changed files and lines, including tests, as in Appendix~\ref{app:dataset_composition}.
For SWE-rebench V2.1 and Scale-SWE, patch statistics cover the 13,238 and 20,168 tasks remaining after excluding repositories related to SWE-bench Pro, HWE-Bench, or our source repositories; trajectory statistics cover solved attempts and include reasoning tokens.
Our reference patches are roughly twice as large in changed lines, and the same solver requires more than twice as many turns and generated tokens to solve them.
In addition, 54.3\% (9,913/18,257) of our synthesized tasks originate from pull requests that modify no test or verification path and would therefore be discarded by pipelines requiring pre-existing PR tests.

\begin{table}[!htb]
    \caption{\textbf{\origent SFT-pool tasks versus SWE-rebench V2.1 and Scale-SWE.} Trajectory statistics are measured under the same controlled reference solver (GLM-5.3 in OpenCode). Entries report median (IQR) unless noted.}
    \label{tab:swerebench_comparison}
    \centering
    \small
    \setlength{\tabcolsep}{4pt}
    \begin{tabular*}{\linewidth}{@{\extracolsep{\fill}}lrrr@{}}
        \toprule
        Metric & \origent SFT pool & SWE-rebench V2.1 & Scale-SWE \\
        \midrule
        Files changed per reference PR (median) & 4 & 3 & 3 \\
        Lines changed per reference PR & 132 (54--325) & 75 (37--153) & 66 (27--176) \\
        Solved solver trajectories (count) & 8,013 & 968$^{\dagger}$ & 415$^{\dagger}$ \\
        Assistant turns per trajectory & 113 (69--166) & 41 (27--69) & 33 (19--49) \\
        Generated tokens per trajectory (K) & 56.8 (32.1--88.9) & 25.8 (10.3--48.2) & 17.7 (5.6--37.6) \\
        \bottomrule
    \end{tabular*}

    \vspace{3pt}
    {\footnotesize $^{\dagger}$From uniform random samples of 1,263 SWE-rebench V2.1 tasks and 759 Scale-SWE tasks.}
\end{table}

\section{Construction Agents and Execution Checks}
\label{app:construction_agents}

\paragraph{Models and harnesses.}
Repository profiles are authored by GPT-5.6 Sol in Codex.
Task construction uses GPT-5.6 Sol in Codex and, for later batches, GLM-5.3 in OpenCode v1.18.22; problem-statement generation and review use GLM-5.3 in separate Codex or OpenCode sessions.

\paragraph{Profile preparation.}
Given a repository's build files and candidate metadata, the profile agent identifies the required compilers, simulators, and libraries, defining multiple profiles whenever historical revisions require different toolchains.
It outputs a repository template (base image, dependencies, setup script, and working directory) together with repo-specific task guidance, from which a Dockerfile is rendered and checked against a small batch of candidate PRs before broad reuse.
When task agents encounter missing shared dependencies, they flag the profile for revision.
Across the 18,257 synthesized tasks, 152 repository-specific profiles were used (mean 120.1 tasks per profile, median 38.5; 53 of the 90 repositories need only a single profile), consolidated into 148 toolchain profiles across 90 repository templates in the open-source codebase.

\paragraph{Task construction and execution checks.}
For each candidate PR, the task agent receives the base commit, reference patch, PR metadata, repository template, and repo-specific guidance.
It selects a profile and writes three scripts: \texttt{prepare\_script.sh} for task-specific setup, \texttt{test.sh} for the target check (creating any temporary testbenches or stimuli needed), and \texttt{regression.sh} for regression coverage, iterating against live build and test output.

Final validation runs four checks in fresh containers created from the same prepared image, applying the reference patch only on the repaired-state runs.
The base-state target check must exit nonzero and print the designated behavioral failure marker (timeouts and build errors are rejected), the repaired-state target check must exit zero with its success marker, and both regression checks must exit zero.
Appendix~\ref{app:task_example} shows an example of the scripts and recorded outcomes.

\paragraph{Passing patches versus reference patches.}
Because synthesized tests target observable behavior rather than private implementation details (Section~\ref{sec:behavioral_synthesis}), valid alternative fixes should pass without copying the reference diff.
Table~\ref{tab:patch_gap} compares passing GLM-5.3 patches against reference patches across 7,758 comparable SFT-pool tasks after restricting diffs to source files and excluding test paths, documentation, build outputs, and binaries.
A reference added line counts as reused if it contains at least four non-whitespace characters and appears verbatim, modulo surrounding whitespace, in the passing patch.
Passing patches typically edit the same source files as the reference while reusing only a small fraction of its added lines.

\begin{table}[!htb]
    \caption{\textbf{Passing solver patches versus reference patches on SFT-pool tasks} (7,758 comparable GLM-5.3 solutions).}
    \label{tab:patch_gap}
    \centering
    \small
    \begin{tabular*}{\linewidth}{@{\extracolsep{\fill}}lr@{}}
        \toprule
        Metric & Value \\
        \midrule
        Reference files also edited, mean & 72.9\% \\
        Reference added lines reused, median (IQR) & 11.7\% (2.8\%--28.6\%) \\
        Tasks reusing less than half of reference added lines & 88.5\% \\
        \bottomrule
    \end{tabular*}
\end{table}

\paragraph{Problem-statement generation and review.}
After execution checks pass, a generation agent writes the problem statement from the original issue or PR, the base code, the tests, and the execution logs, capturing all tested interface and behavioral requirements while omitting reference patch excerpts or repair hints.
A separate review agent independently inspects the source PR and execution evidence, checks the draft for missing requirements or ambiguity, and returns \texttt{approved}, \texttt{revised}, or \texttt{rejected}.
Approved and revised statements are kept; rejected tasks are dropped.
Across 18,080 reviewed statements, 48.3\% were approved as drafted and 51.7\% were revised.

\section{Example of a Constructed Task}
\label{app:task_example}

This task, derived from LUNA PR~113, concerns the constant-stream generator used in USB gateware.
When transmission starts at a nonzero byte offset, the first payload incorrectly comes from offset zero.
The task uses the \texttt{legacy-nmigen-py39} profile, which prepares Python~3.9 and the historical nMigen dependencies.
Both tests simulate the hardware through LUNA's test harness in the USB clock domain.
The excerpts below show the stimuli and checks, omitting test-runner setup and diagnostic logging.

\paragraph{Problem statement.}
Generated task description provided to the agent:
\begin{quote}
\small
ConstantStreamGenerator does not emit the requested byte when it is started with a nonzero start\_position. For example, with constant\_data = b"HELLO, WORLD", start\_position = 2, and a positive max\_length, the first beat after start is asserted currently presents H, the byte at offset 0. Instead, when that first beat is presented, stream.valid and stream.first must be asserted and stream.payload must equal constant\_data[2], i.e. L. More generally, the first payload emitted after a start strobe must be the byte at the supplied start\_position, not the byte at offset 0.

The existing zero-offset behavior must remain unchanged: after start\_position = 0 and a positive max\_length, stream.valid and stream.first are asserted with stream.payload equal to constant\_data[0]; that byte remains presented while stream.ready is low, and stream.payload advances to constant\_data[1] after stream.ready is asserted. This affects the byte-oriented constant-stream feature used for USB descriptor responses, including descriptor data resumed at a nonzero position.

\end{quote}

\paragraph{Target test.}
The test configures \texttt{ConstantStreamGenerator} with \texttt{b"HELLO, WORLD"} and starts transmission at offset~2.
It checks the first payload and its valid/first indicators; the expected byte is \texttt{L}.

\medskip
\noindent\begin{minipage}{\linewidth}
\small
\begin{verbatim}
yield dut.start_position.eq(2)
yield dut.max_length.eq(1000)
yield from self.advance_cycles(4)
yield from self.pulse(dut.start)

valid = yield dut.stream.valid
first = yield dut.stream.first
actual = yield dut.stream.payload
expected = ord("L")

if valid != 1 or first != 1 or actual != expected:
    self.fail(
        f"first valid payload at offset 2: valid={valid}, "
        f"first={first}, expected={expected}, actual={actual}"
    )
\end{verbatim}
\end{minipage}
\medskip

\noindent\begin{minipage}{\linewidth}
\paragraph{Regression test.}
Using the same constant data, the regression test starts at offset~0 and holds \texttt{stream.ready} low for three cycles.
The first byte must remain \texttt{H} until the receiver becomes ready, after which the stream advances to \texttt{E}.

\medskip
\small
\begin{verbatim}
yield dut.start_position.eq(0)
yield dut.max_length.eq(5)
yield dut.stream.ready.eq(0)
yield from self.advance_cycles(4)
yield from self.pulse(dut.start)

self.assertEqual((yield dut.stream.valid), 1)
self.assertEqual((yield dut.stream.first), 1)
self.assertEqual((yield dut.stream.payload), ord("H"))

yield from self.advance_cycles(3)
self.assertEqual((yield dut.stream.valid), 1)
self.assertEqual((yield dut.stream.first), 1)
self.assertEqual((yield dut.stream.payload), ord("H"))

yield dut.stream.ready.eq(1)
yield
yield
self.assertEqual((yield dut.stream.valid), 1)
self.assertEqual((yield dut.stream.first), 0)
self.assertEqual((yield dut.stream.payload), ord("E"))
\end{verbatim}
\end{minipage}
\medskip

Table~\ref{tab:example_validation} reports the recorded outcomes on the base and repaired states.

\begin{table}[!htb]
    \caption{\textbf{Recorded execution checks for LUNA PR~113.}}
    \label{tab:example_validation}
    \centering
    \small
    \begin{tabular*}{\linewidth}{@{\extracolsep{\fill}}llll@{}}
        \toprule
        Repository state & Check & Exit code & Recorded outcome \\
        \midrule
        Base & Target & 1 & First payload: \texttt{H}; expected \texttt{L} \\
        Repaired & Target & 0 & First payload: \texttt{L} \\
        Base & Regression & 0 & Hold \texttt{H}, then advance to \texttt{E} \\
        Repaired & Regression & 0 & Hold \texttt{H}, then advance to \texttt{E} \\
        \bottomrule
    \end{tabular*}
\end{table}

\FloatBarrier
\section{Training Settings and Telemetry}
\label{app:training_evaluation_settings}

\subsection{Training Configuration}
\label{app:training_configuration}

Table~\ref{tab:training_settings} lists the SFT and online GRPO configurations.

\begin{table}[!htb]
    \caption{\textbf{Training settings for SFT and online GRPO.} TP, PP, CP, DP, EP, and ETP denote tensor, pipeline, context, data, expert, and expert tensor parallelism, respectively.}
    \label{tab:training_settings}
    \centering
    \small
    \begin{tabular*}{\linewidth}{@{\extracolsep{\fill}}lll@{}}
        \toprule
        Setting & SFT & RL (GRPO) \\
        \midrule
        H100 80GB GPUs & 64 & 64 \\
        Training TP / PP / CP / DP & 2 / 4 / 4 / 2 & 2 / 4 / 4 / 2 \\
        Training EP / ETP & 8 / 1 & 8 / 1 \\
        Rollout TP / EP & --- & 8 / 8 \\
        Concurrent agent sandboxes & --- & 256 (32 per node) \\
        \midrule
        Training tasks & 8,444 (9,664 segments) & 2,163 \\
        Training repositories & 61 & 23 \\
        Batch size & 32 segments & 64 groups $\times$ 8 rollouts \\
        Duration & 3 epochs (906 updates) & 51 updates; 1/batch \\
        Learning rate & $10^{-5}\rightarrow10^{-6}$ & $10^{-6}$ \\
        Schedule & Cosine; 5\% warmup & Constant; no warmup \\
        Optimizer & Adam ($\beta=0.9, 0.95$) & Adam ($\beta=0.9, 0.999$) \\
        Weight decay / grad clipping & 0.01 / 1.0 & 0.01 / 1.0 \\
        Context limit & 256K (262,144) tokens & 256K (262,144) tokens \\
        Generation limit per response & --- & 64K (65,536) tokens \\
        Sampling temperature / top-$p$ & --- & 1.0 / 1.0 \\
        Advantage normalization & --- & Mean-centered (unscaled) \\
        MoE router weights & Trainable & Frozen \\
        KL / entropy coefficient & --- & 0 / 0 \\
        Compaction reserved / threshold & --- & 72K / 184K tokens \\
        Agent / verifier time limit & --- & 3 hours / 1 hour \\
        \bottomrule
    \end{tabular*}
\end{table}

\paragraph{Optimization and prompt-mean loss aggregation.}
For each prompt group $g \in \{1,\dots,G\}$ ($G=64$) containing $M=8$ trajectories with binary rewards $r_{g,i} \in \{0,1\}$, advantages are mean-centered without standard-deviation scaling:
\begin{equation}
    A_{g,i} = r_{g,i} - \frac{1}{M}\sum_{j=1}^{M} r_{g,j}.
    \label{eq:grpo_advantage}
\end{equation}
Because context compaction may split trajectory $(g,i)$ into $K_{g,i}$ training segments, let $m_{g,i,s,t}\in\{0,1\}$ denote the supervision mask for token $y_{g,i,s,t}$ in segment $s$, and let $N_g = \sum_{i=1}^{M}\sum_{s=1}^{K_{g,i}}\sum_t m_{g,i,s,t}$ be the total supervised token count across all eight trajectories in group $g$.
With a single optimizer step per batch, the baseline log-probability is detached from the same training forward pass ($b_{g,i,s,t} = \operatorname{sg}[\log\pi_\theta(y_{g,i,s,t}\mid c_{g,i,s,t})]$), so $\rho_{g,i,s,t} = \exp(\log\pi_\theta - b_{g,i,s,t})$ evaluates to 1 while gradients flow through the numerator.
We minimize the prompt-mean token loss:
\begin{equation}
    \mathcal{L}_{\mathrm{GRPO}}(\theta) =
    -\frac{1}{G}\sum_{g=1}^{G}
    \frac{1}{N_g}\sum_{i=1}^{M}\sum_{s=1}^{K_{g,i}}\sum_{t}
    m_{g,i,s,t}\,\rho_{g,i,s,t}(\theta)\,A_{g,i}.
    \label{eq:prompt_mean_grpo}
\end{equation}
The mask excludes prompts, tool outputs, and context copies of prior summaries while including admitted summary generations.

\paragraph{Context handling, replay, and stability guards.}
OpenCode reserves 72K tokens and triggers compaction at 184K tokens with tool-output pruning disabled.
A generated summary is supervised only if it comes from native compaction, ends with \texttt{finish=stop}, has nonempty visible text, and is consumed by a subsequent turn.
Miles TITO records exact prompt and generated token IDs during SGLang rollout and passes them unchanged to Megatron training; $R^3$ replays the recorded MoE expert choices per token and layer; and MoE router weights stay frozen throughout RL.
Degenerate outputs are masked asymmetrically across outcomes: length-truncated responses receive zero loss on passing trajectories while keeping negative gradients on failed ones, and long n-gram loops ($n=300$, window 8,300, minimum length 16,384, at least 15 repeats) are masked in full on passing trajectories and after the first repeat trigger on failed ones.
To guard against occasional inference-side NaN logits that collapse sampling to uniform token selection ($\log p \approx -\ln V$), our released dynamic-sampling filter detects contiguous uniform-logprob runs ($\geq 64$ tokens) and drops affected groups for resampling.

\paragraph{Synchronous execution and compute.}
All 51 RL updates on 64 H100 GPUs took 368.6 wall-clock hours ($\approx$23,600 GPU-hours): 85.0\% in rollout (sandboxes, SGLang inference at 96.7\% median prefix-cache hit rate, and verification), 12.4\% in Megatron training (23.7\% median training MFU), and 2.6\% in weight sync and checkpointing.
Each update kept 64 mixed groups while dynamic filtering discarded a median of 21 all-pass ($8/8$), 5 all-fail ($0/8$), and 8 infrastructure-aborted or corrupted groups (median 7.8\% of completed groups).

\subsection{SFT Data and Supervision Statistics}
\label{app:sft_statistics}

On the 8,861 SFT-pool tasks, DeepSeek-V4.1 Flash produced 8,607 passing trajectories; quality filtering retained 8,447 tasks (9,060 interaction segments) across 61 repositories plus 613 reconstructed compaction-summary samples.
Dropping three segments exceeding 262,144 tokens, three all-masked summary samples, and three trailing segments for batch alignment leaves 8,444 tasks and 9,664 training segments (9,054 interaction segments and 610 summary samples).
Each epoch processes 1.07B input tokens, of which 256.3M are supervised across 436,576 assistant turns.
Tables~\ref{tab:sft_turn_masks} and~\ref{tab:sft_lengths} break down turn masking and length quantiles.

\begin{table}[!htb]
    \caption{\textbf{Turn-level SFT reasoning mask.} Counts across all 9,664 final SFT training segments (tool-error and incomplete turns are masked prior to tokenization).}
    \label{tab:sft_turn_masks}
    \centering
    \small
    \begin{tabular*}{\linewidth}{@{\extracolsep{\fill}}lr@{}}
        \toprule
        Category & Assistant turns \\
        \midrule
        Interaction turns before reasoning mask & 475,724 \\
        \quad Empty reasoning (masked) & 39,484 \\
        \quad Reasoning longer than 8,192 tokens (masked) & 274 \\
        Interaction turns retained for supervision & 435,966 \\
        Compaction-summary turns retained for supervision & 610 \\
        \midrule
        Total supervised assistant turns & 436,576 \\
        \bottomrule
    \end{tabular*}
\end{table}

\begin{table}[!htb]
    \caption{\textbf{SFT length distributions.} Per-trajectory and per-segment quantiles across the final SFT training set (8,444 tasks, 9,664 segments).}
    \label{tab:sft_lengths}
    \centering
    \small
    \begin{tabular*}{\linewidth}{@{\extracolsep{\fill}}lrrrrr@{}}
        \toprule
        Quantity & Median & 90th & 95th & 99th & Maximum \\
        \midrule
        Assistant turns per trajectory & 50 & 97 & 120 & 171 & 293 \\
        Input tokens per segment & 102,492 & 198,412 & 216,505 & 229,026 & 261,715 \\
        Supervised tokens per segment & 23,637 & 52,360 & 60,236 & 75,440 & 112,294 \\
        \bottomrule
    \end{tabular*}
\end{table}

\subsection{RL Task Selection}
\label{app:rl_selection_counts}

Table~\ref{tab:rl_screening_counts} summarizes the RL task pool after Qwen3.6-27B screening across 23 disjoint repositories.
Tasks in the $0/4$ bin are kept only if GLM-5.3 solves them in the same prepared container image.

\begin{table}[!htb]
    \caption{\textbf{RL task screening.} Candidate and retained counts across 23 disjoint repositories by Qwen3.6-27B screening pass rate ($0/4$, $1/4$, and $2/4$).}
    \label{tab:rl_screening_counts}
    \centering
    \small
    \begin{tabular*}{\linewidth}{@{\extracolsep{\fill}}lrrrrr@{}}
        \toprule
        Repositories & Candidates & $0/4$ & $1/4$ & $2/4$ & Retained \\
        \midrule
        23 & 8,199 & 705 & 649 & 809 & 2,163 \\
        \bottomrule
    \end{tabular*}
\end{table}

\subsection{Training Stability and Rollout Telemetry}
\label{app:train_rollout_consistency}

\begin{figure}[!htb]
    \centering
    \includegraphics[width=\linewidth]{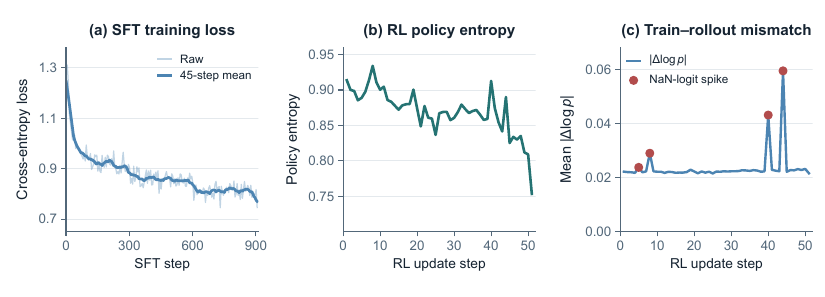}
    \caption{\textbf{Training stability telemetry.} \textbf{(a)}~SFT cross-entropy loss over 906 steps. \textbf{(b)}~RL policy entropy over 51 updates. \textbf{(c)}~Mean absolute train--rollout log-probability difference across 51 updates; red dots mark the four steps with rare inference-side NaN-logit trajectories.}
    \label{fig:rl_stability}
\end{figure}

Figure~\ref{fig:rl_stability} and Table~\ref{tab:rl_rollout_behavior} report training stability and rollout behavior across SFT and online GRPO.
Across 906 SFT updates, cross-entropy loss drops from 1.29 to 0.75 (Figure~\ref{fig:rl_stability}a).
Over 51 GRPO updates, policy entropy falls from 0.914 to 0.753 (Figure~\ref{fig:rl_stability}b).
With TITO, $R^3$, and frozen MoE routers enabled, train--rollout mismatch stays in a narrow band on normal steps (Figure~\ref{fig:rl_stability}c): mean $|\Delta\log p|$ lies in $[0.0215, 0.0232]$ (median 0.0222), steady-state KL has median 0.0012, and a median of 0.63\% of tokens fall outside a $[0.8, 1.25]$ probability ratio.
The four marked spikes (steps 5, 8, 40, and 44) stem from rare SGLang NaN logits causing uniform sampling in a few trajectories; when the same issue recurred in four groups at the final step~51, we dropped those groups and reran step~51 from the step-50 checkpoint, and enabled the uniform-sampling filter by default in the released code.

\begin{table}[!htb]
    \caption{\textbf{RL rollout trajectory statistics.} Per-trajectory behavior across retained mixed-outcome training batches (64 groups $\times$ 8 rollouts = 512 trajectories per update), averaged over the first five updates (steps 1--5) versus the last five updates (steps 47--51). Assistant turns exclude context-compaction summaries.}
    \label{tab:rl_rollout_behavior}
    \centering
    \small
    \begin{tabular*}{\linewidth}{@{\extracolsep{\fill}}lrrr@{}}
        \toprule
        Metric & Steps 1--5 & Steps 47--51 & Change \\
        \midrule
        Assistant turns per trajectory (mean) & 240.8 & 293.2 & +21.8\% \\
        Tool calls per trajectory (mean) & 279.3 & 368.1 & +31.8\% \\
        Cumulative output tokens (mean, K) & 148.6 & 205.3 & +38.2\% \\
        Context compactions per trajectory (mean) & 1.86 & 3.08 & +65.6\% \\
        3-hour timeout rate (\%) & 11.1 & 18.7 & +7.6 pp \\
        \bottomrule
    \end{tabular*}
\end{table}

\end{document}